\documentclass[letterpaper, 10 pt, journal, twoside]{IEEEtran}

\usepackage{color}
\usepackage{graphicx}
\usepackage{float}
\usepackage{amsmath}

\usepackage{soul}
\usepackage{verbatim}

\definecolor{Gray}{gray}{0.9}

\title{\LARGE \bf
Integrating Elastic Bands to Enhance Performance for Textile Robotics
}

\author{Cem Suulker$^{1}$, Sophie Skach$^{1}$, and Kaspar Althoefer$^{1}$%

\thanks{$^{1}$All authors are with the Centre for Advanced Robotics, at the School of Engineering and Materials Science, Queen Mary University of London, United Kingdom.
        {\tt\footnotesize c.suulker@qmul.ac.uk}}%

}

\begin{document}

{

}

\maketitle




\section{Introduction}


The field of soft robotics is growing rapidly, and innovative elastic materials are replacing heavy metallic links while soft inflatable actuators are taking the place of electromechanical rotational motors. The use of flexible textile materials in human-robot interaction has also been shown to offer attractive design options due to their safe nature \cite{cappello2018assisting,suulkercomparison}.


The creation of soft robots and actuators often involves the use of various materials and methods from the clothing industry. One crucial variable is the stretch quality of the fabric material. Knitted fabrics are commonly used for their stretchiness, but they often stretch in all directions, which is usually not desirable. Woven fabrics, are normally known as non-stretchy. But integrating elastane yarn into the weft, makes them more stretch in one direction than the other. This quality makes them more suitable for creating actuators. They are also more durable than knitted ones. While using coating can make these structures airtight, it also eliminates the material's ability to stretch.


Another approach to creating stretch in soft robotic structures involves increasing the material density using methods such as pleats and ruffles. The use of pleats in soft robotics has been extensively researched and is considered an effective method \cite{cappello2018assisting}. However, the use of ruffles has not received as much attention. Ruffles are often used in the neck area of clothing to gather the fabric material in a dense way. From a soft robotics perspective, this gathered dense material can be inflated to create high elongation.


Various types of elastic bands can be used to create ruffles, but in this extended abstract, we will focus on braided elastic bands. These bands are used for storing energy in soft robotics. However, their full potential is realized when they are integrated into the fabric structure using the ruffles method. In this extended abstract, we will showcase two soft robotic applications that utilize elastic bands to enhance the system's performance.



\begin{figure}[t]
  \centering
  \includegraphics[width=1\linewidth]{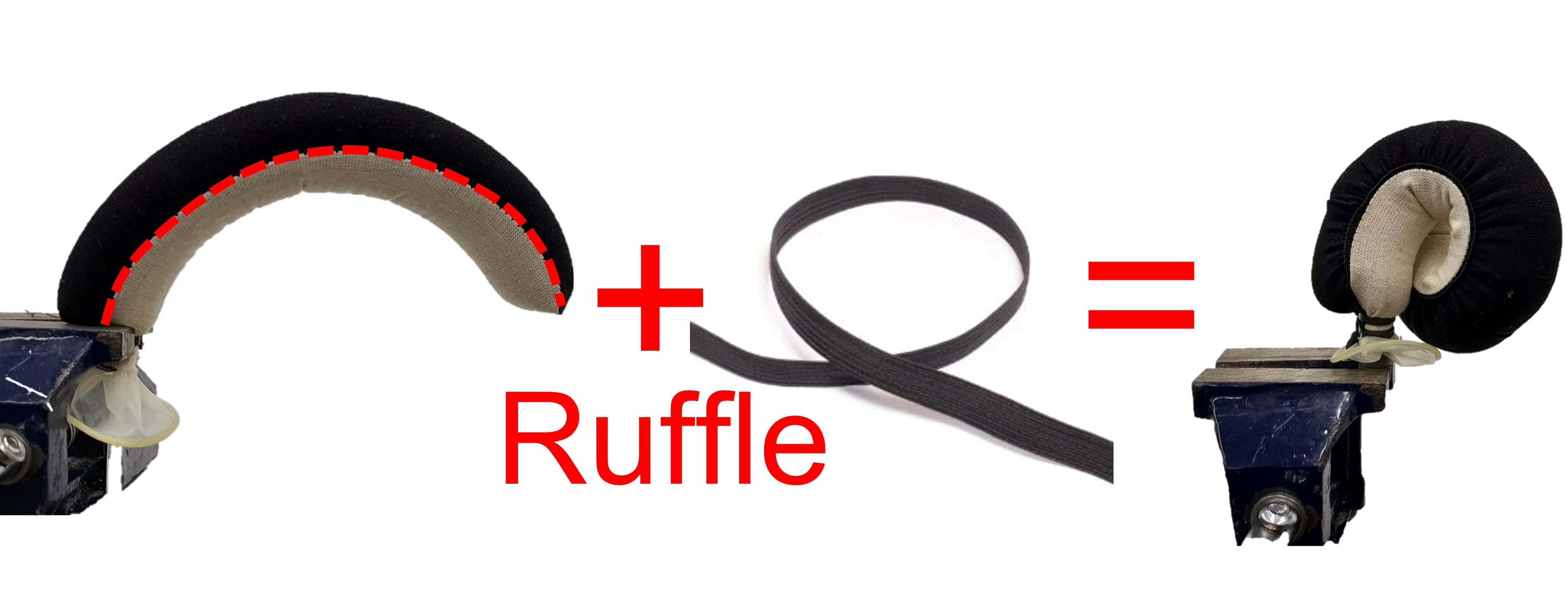}
  \caption{Before and after the integration of elastic bands to an actuator.}
  \label{fig1}
\end{figure}

\section{Application: Bending Actuator for Wearables} 


To create bending of an inflatable textile actuator, an imbalance between two layers of fabric must be achieved. This imbalance is created by using pleating techniques \cite{cappello2018assisting}, excess material of one layer is folded and stitched onto the other, and unfolds when inflated. This approach, however, has potential drawbacks, e.g. when the space between the layers is too small for the pleats to unfold (which, for finger sized designs, can become imminent). Another technique to create such imbalance is the use of different types of fabric with varying elasticity. This way, one layer stretches more when actuated and the structure will bend towards the less elastic one. Integrating braided elastic band to the stretch fabric option with the ruffles technique is proven more efficient in terms of blocking force and bending angle capabilities \cite{suulkeractuator}.



\subsubsection{Materials} 

To create this textile actuator the two different textile properties layers are selected. Bottom: a plain cotton weave for the (light fabric in Fig. 1). Top: a cotton mix with elastane yarn integrated to the weft of the fabric (dark fabric in Fig. 1).


To enhance a fabric's stretch behavior and create the desired material imbalance between layers, an additional support material is used that is integrated when assembling the actuators: a braided elastic band, also called elastics, see Figure \ref{fig1}, commonly used in clothing to create ruffles or elastic waistbands. It consists of braided polyester and a small part of thin rubber, making it durable and extremely stretchy.


\subsubsection{Fabrication}



First, the mono-directional stretch black fabric was cut 80\% longer than the cotton bottom layer fabric. The elastic band was integrated on the side seam between the top and bottom layer, first stitched onto the top layer while being stretched, and then sewn onto the bottom layer in a relaxed state. This enabled the top layer to reach an excess length of 180\% of the bottom layer. The actuator is equipped with a latex bladder to ensure air tightness.

\begin{figure}[h]
\centering
  \includegraphics[width=1\linewidth]{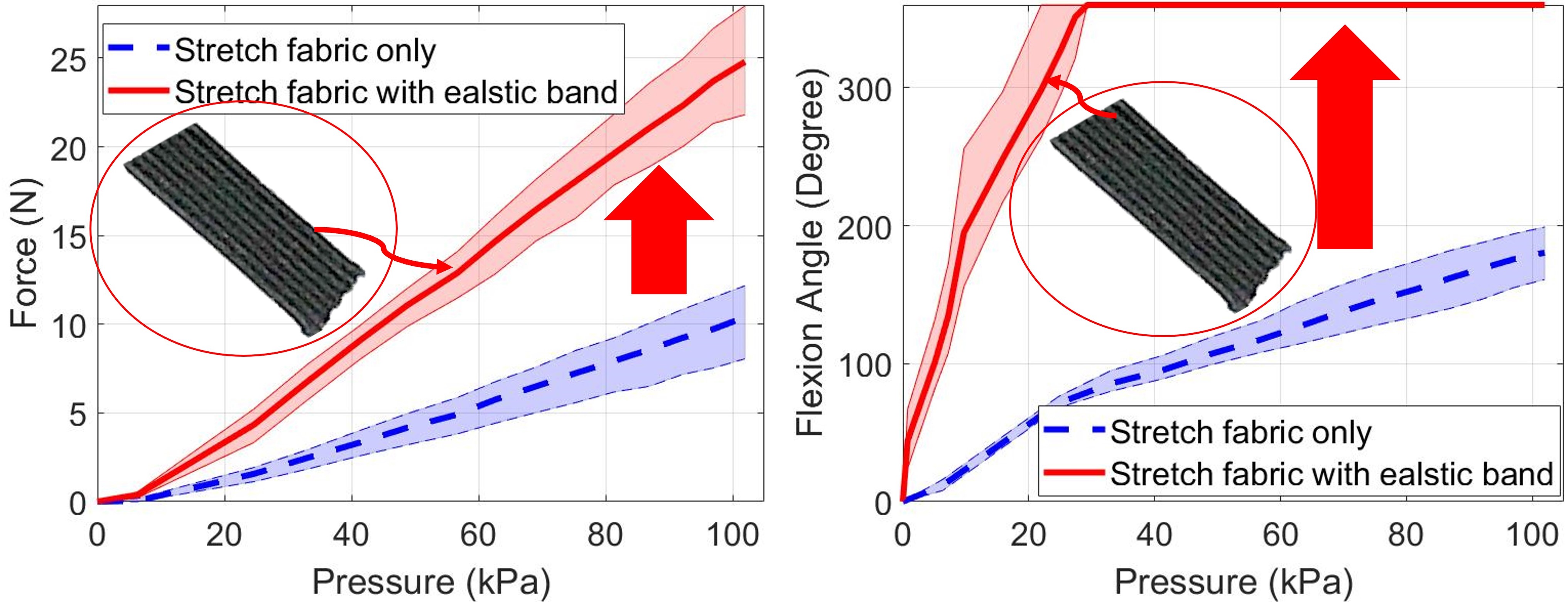}
  \caption{Blocking force output and flexion angle versus pressure graphs for stretch fabric actuator, and elastic band integrated actuator. Integration of the elastic band significantly boosts the performance.}
  \label{fig2}
\end{figure}

\subsubsection{Results}
Two important parameters for soft bending actuators for wearables are flexion angle and blocking force \cite{suulkerexos} In example, for rehabilitation or assistive hand exoskeletons it is imperative that each finger is unrestricted in relation to its maximum angle, and actuators should apply 10-15 N blocking force to the fingers\cite{takahashi2008robot}. 

In Fig. 2 the elastic band integration boosts both of these critical parameters for the actuator. The force capability increases approximately from 10 N to 25 N, and maximum bending angle increases from approximately 180 degrees to 360 degrees.

\section{Application: Soft Cap for Eversion Robots}

Growing robots based on the eversion principle are known for their ability to extend from within rapidly, along their longitudinal axis, and in doing so, reach deep into hard-to-access, remote spaces. Because of their unique movement principle maintaining a payload at the tip is a major challenge. Various tip mechanisms have been proposed, including complex, rigid designs that may not be compatible with functional hardware. To address these shortcomings, we proposed a soft, entirely fabric-based cylindrical cap that can be easily slipped onto the tip of eversion robots \cite{suulkercap}.

We created a series of caps of different sizes and materials and conducted an experimental study to evaluate their effectiveness in maintaining their position and transporting payloads, such as a camera, across long distances. We also assessed the caps' ability to navigate through narrow openings and found that our soft, flexible cap does not significantly hinder the robot's flexibility or overall maneuverability. Our design offers a solution to the challenge of maintaining sensory payloads at the tip of eversion robots without compromising their performance or flexibility.

\begin{figure}[h]
  \centering
  \includegraphics[width=0.7\linewidth]{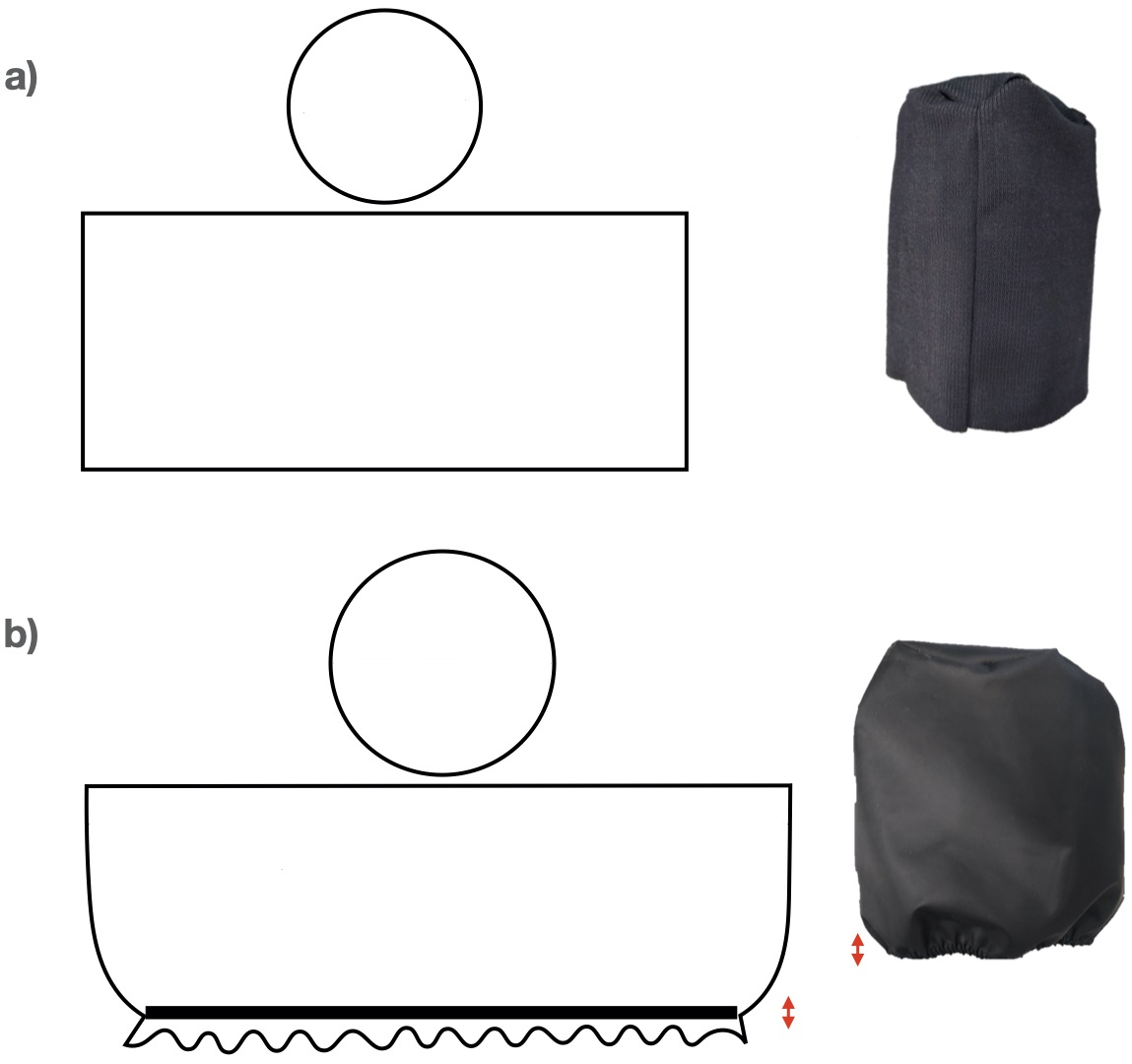}
  \caption{a) Pattern for stretch fabric cap b) Pattern for elastic band integrated cap.}
  \label{fig3}
\end{figure}

\subsubsection{Materials \& Fabrication} 
To keep the cap attached to the tip of the eversion body, the main concept is based on squeezing the cap against the body. We have devised two different designs to achieve this squeezing motion: one using a stretch fabric and the other utilizing ruffles with elastic bands. Please see Fig. \ref{fig3} for more information on the patterns of the caps.

\subsubsection{Results}

In order to assess the performance of the caps, we subjected them to a series of challenges that evaluated their ability to adapt to varying layer thicknesses, protruding objects from the robot body, navigability, and squeezability. The performance of the caps was measured using a percentage index that took into account the number of challenges that the caps successfully completed.

\begin{figure}[h]
  \centering
  \includegraphics[width=0.9\linewidth]{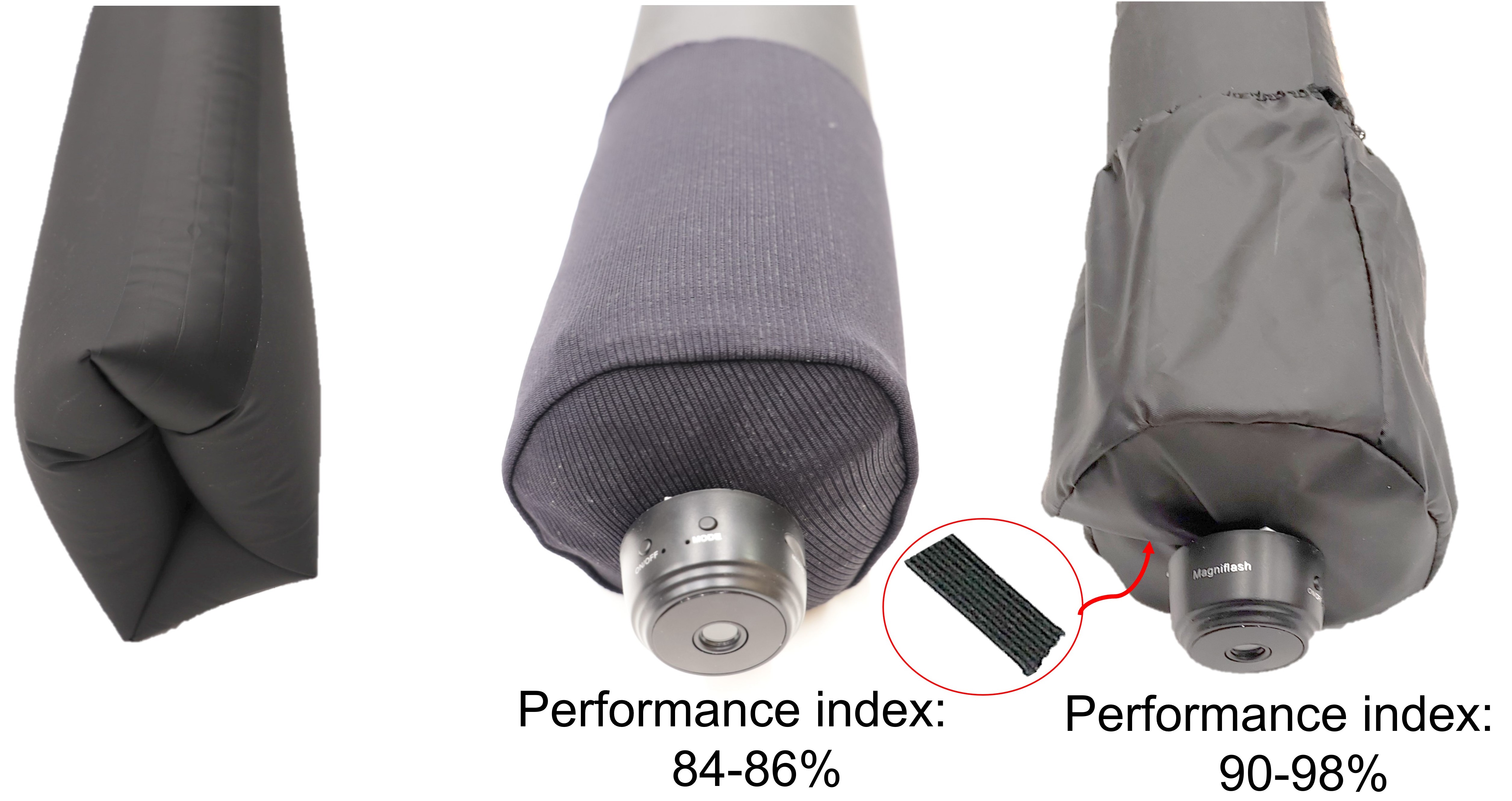}
  \caption{The eversion robot, with a stretchy cap, with an elastic band integrated cap. Integration of elastic band boosts the performance of the robot.}
  \label{fig4}
\end{figure}

The stretch fabric caps achieved a performance rating of 84-86\%, while the elastic band integrated caps achieved a rating of 90-96\% (Fig. \ref{fig4}). This improvement can be attributed to the smaller contact area between the two fabrics, with the stretch fabric caps having a contact area of 15 cm and the elastic band integrated caps having a contact area of only 1 cm.

\section{Conclusions}


 The elastic bands integrated using the ruffles technique proved to be effective in enhancing the performance of the soft robotic structures. In the actuator application, the elastic bands greatly increased the bending capability and force capability of the structure, while in the eversion robot cap application, the elastic bands improved the performance slightly by maintaining the sensory payload at the tip without restricting the eversion process. These findings demonstrate the potential of using elastic bands and textile techniques in soft robotics to create more efficient and adaptable structures.

\addtolength{\textheight}{-12cm}   








 \bibliographystyle{IEEEtran} 

\bibliography{references2}

\end{document}